\documentclass[letterpaper, 10pt, journal, twoside]{IEEEtran}
\usepackage{cite}
\usepackage{amsmath,amssymb,amsfonts}
\usepackage{mathtools}
\usepackage{dsfont}
\usepackage{graphicx}
\usepackage{textcomp}
\usepackage{booktabs}
\usepackage{siunitx}
\usepackage{stfloats}
\usepackage{url}
\usepackage{array}
\usepackage{makecell}
\usepackage{tabularx}
\newcolumntype{Y}{>{\centering\arraybackslash}X}
\usepackage[T1]{fontenc}
\usepackage{microtype}

\usepackage[table]{xcolor}
\definecolor{applegray}{RGB}{142,142,147}
\definecolor{capL}{HTML}{FDF2E4}
\definecolor{capM}{HTML}{F6C98E}
\definecolor{capH}{HTML}{E88A2B}
\definecolor{bestopen}{HTML}{E4F2DC}
\definecolor{bestcomm}{HTML}{E2ECF7}
\definecolor{ourrow}{HTML}{FFF3CC}

\usepackage{hyperref}
\hypersetup{hidelinks}

\newcommand{\swatch}[1]{{\setlength{\fboxsep}{0pt}\setlength{\fboxrule}{0.3pt}\fcolorbox{applegray}{#1}{\rule{0pt}{0.7em}\hspace{0.9em}}}}
\newcommand{\papertablestyle}{%
  \scriptsize
  \setlength{\tabcolsep}{2.5pt}%
  \renewcommand{\arraystretch}{1.05}%
  \setlength{\heavyrulewidth}{0.6pt}%
  \setlength{\lightrulewidth}{0.4pt}%
  \setlength{\cmidrulewidth}{0.4pt}%
}

\AtBeginDocument{%
}

\AtBeginDocument{%
  \setlength{\abovedisplayskip}{3pt plus 1pt minus 1pt}%
  \setlength{\belowdisplayskip}{3pt plus 1pt minus 1pt}%
  \setlength{\abovedisplayshortskip}{2pt plus 1pt}%
  \setlength{\belowdisplayshortskip}{3pt plus 1pt}%
  \setlength{\jot}{3pt}%
}

\makeatletter
\def\section{\@startsection{section}{1}{\z@}{0.8ex plus 0.4ex minus 0.3ex}%
{0.5ex plus 0.5ex}{\normalfont\normalsize\centering\scshape}}%
\def\subsection{\@startsection{subsection}{2}{\z@}{0.8ex plus 0.4ex minus 0.3ex}%
{0.5ex plus 0.3ex}{\normalfont\normalsize\itshape}}%
\makeatother
\let\origthebibliography\thebibliography
\renewcommand{\thebibliography}[1]{\origthebibliography{#1}\fontsize{7.6}{7.7}\selectfont\raggedright}
\NewDocumentCommand\bbm{}{ \begin{bmatrix} }
\NewDocumentCommand\ebm{}{ \end{bmatrix} }
\NewDocumentCommand\Vector{m}{ \boldsymbol{\mathbf{#1}} }
\NewDocumentCommand\Matrix{m}{ \boldsymbol{\mathbf{#1}} }

\begin{document}
\bstctlcite{IEEEexample:BSTcontrol}

\title{MABEL: An Open Anthropomorphic Mobile Bimanual Platform for Teleoperation and Learning}

\author{%
\IEEEauthorblockN{Qilong Cheng$^{1,2,\dagger}$, Rooholla Khorrambakht$^{2}$, Mohamed Yassine Kabouri$^{2,3}$, Quang Nam Nguyen$^{2}$, \\ Luca Macesanu$^{2}$, Haoran Zhou$^{2}$, Yipeng Wang$^{2}$, Ludovic Righetti$^{2}$} \\
\IEEEauthorblockA{\small $^{1}$University of Toronto, Toronto, ON, Canada\\
$^{2}$Center for Robotics and Embodied Intelligence (CREO), New York University, New York, NY, USA\\
$^{3}$LAAS-CNRS, Universit\'e de Toulouse, CNRS, Toulouse, France}
\thanks{$^{\dagger}$Corresponding author: \mbox{qilong.cheng@mail.utoronto.ca}. This work was performed while Q.~Cheng was at New York University and completed at the University of Toronto.}%
}

\markboth{Preprint.}
{Cheng \MakeLowercase{\textit{et al.}}: MABEL: An Open Anthropomorphic Mobile Bimanual Platform}

\IEEEaftertitletext{\vspace{-2.8\baselineskip}}
\maketitle

\begin{abstract}
As robot learning moves beyond the tabletop, many tasks require mobility, adjustable reach, active perception, and bimanual dexterity.
Commercial platforms combining these capabilities are costly, while affordable open-source alternatives cover only a subset and require software integration.
We present \textbf{MABEL}, an open-source anthropomorphic mobile bimanual robot with 56 degrees of freedom (DOF) and a US\$8.1k parts cost.
It combines two 7-DOF arms with 17-DOF hands, a pitching torso, a lift, a 3-DOF neck, and a holonomic base.
Back-drivable arms and torso support compliant interaction, with fingertip reach from floor level to \SI{2.23}{m}.
The head and torso roughly triple modeled camera coverage relative to a torso-fixed camera.
Apple Vision Pro head and hand tracking enables whole-body teleoperation without handheld controllers, with \SI{137}{ms} mean glass-to-glass stereo video latency over the robot's Wi-Fi hotspot.
We demonstrate an end-to-end imitation learning pipeline, from teleoperated data collection to policy deployment, using a shared command interface and whole-body controller.
\end{abstract}


\section{Introduction}
\label{sec:intro}
Recent advances in learning from demonstrations~\cite{aloha2023,diffusionpolicy2023,mobilealoha2024,pi0_2024} have enabled increasingly complex household tasks involving mobile and bimanual dexterous manipulation.
Moving such learning off the tabletop now depends as much on capable embodiments as on better policies.
Collecting the data requires a robot with enough degrees of freedom (DOF) to move, reach across heights, perceive actively, and often manipulate dexterously.
Its actuators must also provide the torque to carry payloads and the back-drivability or torque feedback needed for compliant, contact-rich interaction.
Commercial platforms do offer robots that combine these capabilities, but at the cost of US\$69k--\$120k~\cite{rainbowrobotics,astribot,dexmate}, a substantial barrier to adoption for most research laboratories.
Open-source platforms cost far less, but they often give up capability.
Many rely on position-controlled servos and 1-DOF grippers~\cite{aloha2023,mobilealoha2024,xlerobot,aharobot}, some trade mobility for stronger arms on a fixed base~\cite{openpyro}, and few offer active gaze or adjustable height~\cite{reachy2,yor}.
Many labs instead assemble custom systems from commercial arms, mobile bases, and lifts, leaving each to integrate its own teleoperation, simulation, and data-recording software.

\begin{figure}[t]
    \centering
    \includegraphics[width=\linewidth]{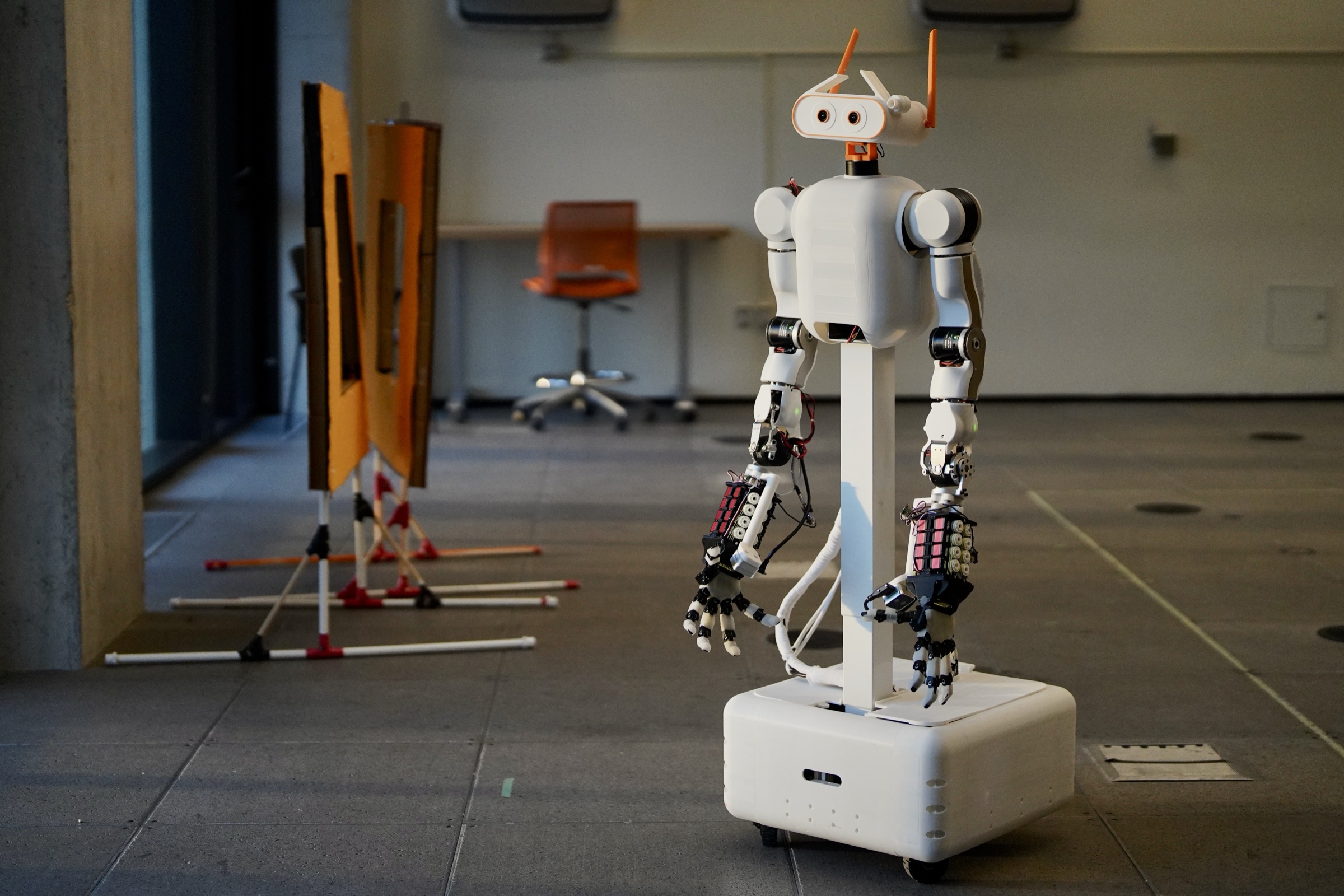}
    \caption{MABEL, a 56-DOF open-source anthropomorphic mobile bimanual robot with a parts cost of US\$8.1k. Hardware designs, firmware, control software, simulation models, and the learning pipeline are available at \href{https://robotmabel.github.io/website/}{robotmabel.github.io/website}.}
    \label{fig:robot_photo}
\end{figure}

\begin{figure*}[t]
\centering
\begin{minipage}[t]{0.62\textwidth}
\vspace{0pt}
\papertablestyle
\setlength{\tabcolsep}{1.7pt}
\begin{tabularx}{\linewidth}{@{}l *{7}{Y}@{}}
\toprule
\rowcolor{applegray!20}
\textbf{Platform} & \textbf{Base} & \textbf{Arms} & \textbf{Hand} & \textbf{Neck}
& \textbf{Body} & \makecell{\textbf{Load}\\\textbf{kg/arm}} & \makecell{\textbf{Cost}\\\textbf{(USD)}} \\
\midrule
\multicolumn{8}{c}{\itshape Open-source platforms}\\
Mobile ALOHA~\cite{mobilealoha2024} & Nonholo. & $6{\times}2$ & 1 & 0 & -- & 0.75 & 32,000 \\
AhaRobot~\cite{aharobot} & Nonholo. & $6{\times}2$ & 1 & \cellcolor{capL}2 & \cellcolor{capL}Lift & 1.5 & 1,000$^{e}$ \\
XLeRobot~\cite{xlerobot} & \cellcolor{capL}Holo. & $5{\times}2$ & 1 & 0 & -- & -- & 660 \\
YOR~\cite{yor} & \cellcolor{capL}Holo. & $6{\times}2$ & 1 & 0 & \cellcolor{capL}Lift & -- & 9,300 \\
TidyBot++~\cite{tidybot2024} & \cellcolor{capL}Holo. & $7{\times}1$ & 1 & 0 & -- & -- & 6,000$^{a}$ \\
OpenPyRo-A1~\cite{openpyro} & Fixed & \cellcolor{capL}$7{\times}2$ & 1 & 0 & \cellcolor{capL}Waist & 5 & 14,000 \\
\rowcolor{bestopen}
Reachy~2~\cite{reachy2} & \cellcolor{capL}Holo. & \cellcolor{capL}$7{\times}2$ & 1 & \cellcolor{capH}3 & -- & 3 & 70,000 \\

\multicolumn{8}{c}{\itshape Commercial platforms}\\
Rainbow RB-Y1~\cite{rainbowrobotics} & Nonholo. & \cellcolor{capL}$7{\times}2$ & 1 & \cellcolor{capL}2 & \cellcolor{capL}Both & 3 & 80,000$^{r}$ \\
Galaxea R1~Pro~\cite{galaxea} & \cellcolor{capL}Holo. & \cellcolor{capL}$7{\times}2$ & 1 & 0 & \cellcolor{capL}Both & 3.5 & 70,000 \\
Astribot S1~\cite{astribot} & \cellcolor{capL}Holo. & \cellcolor{capL}$7{\times}2$ & 1 & \cellcolor{capL}2 & \cellcolor{capL}Both & 5 & 69,000 \\
\rowcolor{bestcomm}
DexMate Vega~\cite{dexmate} & \cellcolor{capL}Holo. & \cellcolor{capL}$7{\times}2$ & \cellcolor{capM}6 & \cellcolor{capH}3 & \cellcolor{capL}Both & -- & 98,000$^{x}$ \\
\midrule
\rowcolor{ourrow}
\textbf{MABEL (ours)} & \textbf{Holo.} & $\mathbf{7{\times}2}$ & \textbf{17} & \textbf{3} & \textbf{Both} & \textbf{3.5} & \textbf{8,056}$^{b}$ \\
\bottomrule
\end{tabularx}
\end{minipage}
\hfill
\begin{minipage}[t]{0.36\textwidth}
\vspace{0pt}
\centering
\includegraphics[width=\linewidth]{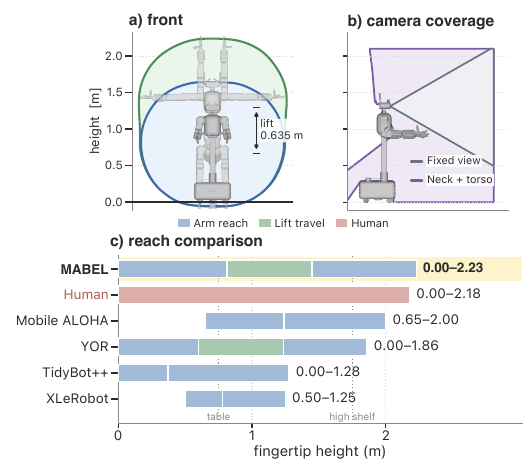}
\end{minipage}

\vspace{1pt}
\parbox{\textwidth}{\scriptsize\noindent
Base: Holo.\ = holonomic, Nonholo.\ = nonholonomic; arms: DOF$\times$count; hand, neck: DOF; body: lift, torso, or both.
Shading: \swatch{capL}~capability shared with MABEL, darkening for higher-DOF hands (\swatch{capM}) and necks (\swatch{capH}); rows: \swatch{bestopen}~most capable open-source and \swatch{bestcomm}~most capable commercial baseline, \swatch{ourrow}~MABEL.
Costs (Oct.\ 2026) are list or reported prices, not scope-matched:
$^{a}$base only;
$^{e}$compute extra;
$^{r}$research edition, \$120,000 commercial;
$^{x}$with hands, \$83,000 without;
$^{b}$Pi~5 parts subtotal, exclusions in \autoref{sec:replication}.}
\vspace{-4pt}
\caption{Platform capabilities, cost, reach, and camera coverage.
MABEL combines a holonomic mobile base, two 7-DOF arms, 17-DOF hands, a 3-DOF neck, a lift, and a pitching torso at roughly a tenth of the cost of comparable commercial platforms.
(a)~Kinematic fingertip workspace. (b)~Modeled fixed-camera view and cumulative neck/torso visibility (\ang{60} vertical field of view (FOV)). (c)~Fingertip-height ranges against other platforms and a human reference, from joint limits only.}
\label{fig:capability_cost_reach}
\end{figure*}

\begin{figure*}[t]
\centering
\includegraphics[width=\textwidth]{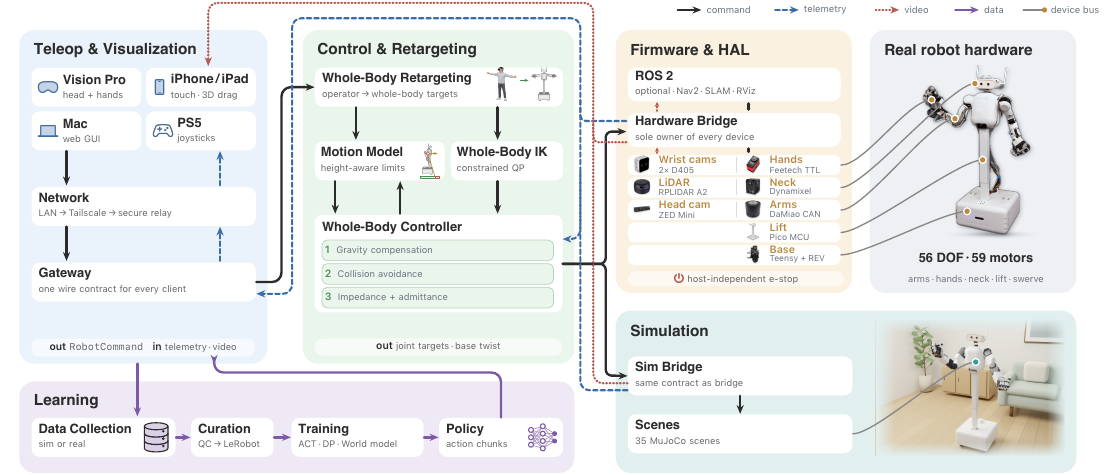}
\caption{MABEL software stack. Teleoperation clients, simulation, and learned policies share one command interface and whole-body controller, which drives either the real robot or its digital twin. HAL: hardware abstraction layer; QP: quadratic program.}
\label{fig:system_architecture}
\end{figure*}

We present \textbf{MABEL}, a Mobile Anthropomorphic Bimanual Embodiment for Learning: an open-source, full-stack platform that meets these requirements at a parts cost of US\$8{,}056 (\autoref{fig:robot_photo}).
Back-drivable 7-DOF arms adapted from OpenArm~\cite{openarm} and a pitching torso support compliant contact.
The torso and a lift extend fingertip reach from the floor to \SI{2.23}{m}, a 3-DOF neck provides active perception, and a three-module swerve base provides omnidirectional mobility.
The end-effector is modular and can be swapped with 17-DOF ORCA hands~\cite{orcahand} or a gripper depending on the research needs (\autoref{tab:specifications}, \autoref{fig:capability_cost_reach}).
With an Apple Vision Pro, the robot becomes the operator's avatar: head and hand tracking alone is retargeted into whole-body motion, and the robot's stereo view returns to the headset at \SI{137}{ms} glass-to-glass latency (\autoref{sec:teleop}).
Teleoperation, simulation, and learned policies share one command interface and whole-body controller, so demonstrations are recorded through the control path that later executes the policy (\autoref{fig:system_architecture}, \autoref{sec:autonomy}).
On hardware, we characterize repeatability, odometry, and latency, demonstrate teleoperated household tasks including compliant interaction, and train and deploy a whole-body Action Chunking with Transformers (ACT) policy to validate the learning pipeline (\autoref{sec:experiments}).
All designs, software, and the digital twin are released, and the robot goes from parts to running in about two weeks (\autoref{sec:mechatronic}).
Our contributions are:
\begin{enumerate}
    \item \textbf{Affordable open hardware:} a US\$8.1k fully open-source mobile bimanual platform integrating compliant actuation, dexterity, mobility, active perception, and floor-to-overhead reach.
    \item \textbf{Whole-body teleoperation:} a controller-free interface that retargets sparse head and hand tracking into whole-body motion with stereo feedback.
    \item \textbf{Unified control and learning stack:} one compliant whole-body controller shared by teleoperation, simulation, data collection, and learned policies, validated end to end with an ACT policy.
\end{enumerate}

\section{Related Work}
\label{sec:related}
\textbf{Mobile bimanual platforms.}
Open-source platforms have progressively added the embodiment that demonstration collection needs (\autoref{fig:capability_cost_reach}).
ALOHA established low-cost bimanual teleoperation on a fixed base~\cite{aloha2023}.
Mobile ALOHA added a nonholonomic base, but the arms remain position-controlled servos with \SI{0.75}{kg} payload, and the workspace is fixed at table height~\cite{mobilealoha2024}.
XLeRobot pushes cost below US\$1k with an omnidirectional base, at the price of 5-DOF servo arms with limited reach and no torso or lift~\cite{xlerobot}.
TidyBot++ introduced a holonomic base carrying a single commercial arm~\cite{tidybot2024}, and YOR extended it with a second arm and a vertical lift~\cite{yor}; both mount the overhead camera rigidly.
AhaRobot pairs a lift with a 2-DOF neck, but on selective compliance articulated robot arm (SCARA)-style servo arms and a nonholonomic base~\cite{aharobot}.
OpenPyRo-A1 moves to 7-DOF torque-capable arms and a waist, with \SI{5}{kg} payload, but on a fixed base and with harmonic-drive actuators that account for much of its cost~\cite{openpyro}.
Reachy~2 combines 7-DOF arms, an active head, and an omnidirectional base in an open design, but without a lift or torso and at about US\$70k~\cite{reachy2}.
Across these, nearly every platform ends in a 1-DOF gripper, with dexterous hands appearing mainly on fixed bases or humanoids~\cite{leaphand2023,opentelevision2024}.
Released simulation also ranges from partial, with YOR welding the base and lift in its runner~\cite{yor} and AhaRobot and OpenPyRo-A1 releasing Unified Robot Description Format (URDF) files without a working simulation~\cite{aharobot,openpyro}, to complete, with TidyBot++ closing the policy-learning loop in MuJoCo~\cite{tidybot2024}.
Commercial platforms supply the combination, pairing 7-DOF arms with torso motion and in some cases active gaze and multi-fingered hands~\cite{rainbowrobotics,astribot,galaxea,dexmate}, but at the cost of US\$69k to over US\$120k.
In addition, without published design files, modification and repair depend on the vendor, and a replacement part can take months to arrive.

MABEL brings together what each of these lacks, back-drivable 7-DOF arms and torso, 17-DOF hands, a 3-DOF active neck, a lift, and an omnidirectional base, for under US\$8.1k, roughly a tenth of the commercial cost.
The computer-aided design (CAD) files, firmware, Robot Operating System~2 (ROS~2) stack, whole-body controller, retargeter, learning pipeline, and a MuJoCo and Isaac Lab digital twin covering the full body are all released (\autoref{sec:mechatronic}).

\textbf{Whole-body teleoperation interfaces.}
A capable embodiment is only useful if an operator can command it naturally and intuitively.
Existing interfaces limit which parts of the robot the operator can command.
Leader--follower arms~\cite{aloha2023,gello2024} and phone or pedal controls leave base, height, and gaze as separately operated channels~\cite{mobilealoha2024,tidybot2024,aharobot}.
JoyLo extends leader arms with joysticks for whole-body control of a wheeled humanoid, but still routes base and torso through handheld input~\cite{brs2025}.
Immersive systems close part of this gap.
Vision-based methods map tracked hands onto dexterous manipulators~\cite{dexpilot2020,anyteleop2023}.
Open-TeleVision and Bunny-VisionPro return stereo video so the operator sees through the robot's cameras~\cite{opentelevision2024,bunnyvisionpro2024}.
Mobile-TeleVision and TeleMoMa further extend the interface to a mobile base~\cite{mobiletelevision2024,telemoma2024}.
Humanoid systems go further, reconstructing whole-body motion from sparse headset and controller poses~\cite{omnih2o2024,hover2025} or full-body tracking~\cite{humanplus2024}, with X-OP adding predictive retargeting and localization feedback~\cite{xop2026}.
Together these show that sparse tracking can drive a high-DOF robot, while standardized data formats and policy architectures determine what the collected demonstrations can train~\cite{droid2024,umi2024,diffusionpolicy2023,pi0_2024}.
MABEL builds on this line of work with a retargeter and motion limits designed around a body proportioned to the operator, so that head and hand tracking alone drives base, lift, torso, neck, arms, and fingers as one body, with on-screen joysticks for the base and lift when the operator's space is tight (\autoref{sec:control}).

\section{Hardware Design}
\label{sec:mechatronic}\label{sec:integration}
\begin{table}[t]
\centering
\caption{MABEL specifications.}
\papertablestyle
\label{tab:specifications}
\begin{tabular}{@{}ll>{\hspace{3pt}}ll@{}}
\toprule
\rowcolor{applegray!20}[\tabcolsep][\tabcolsep]
\textbf{Parameter} & \textbf{Value} & \textbf{Parameter} & \textbf{Value} \\
\midrule
Functional DOF   & 56 (59 actuators)        & Mass             & \SI{61.8}{kg} \\
Overall height   & \SIrange{1.16}{1.80}{m}  & Chassis          & $0.49\times\SI{0.49}{m}$ \\
Support polygon & $0.34\times\SI{0.35}{m}$ & Model load limit & $\sim$\SI{3.5}{kg} \\
Parts subtotal & \$8{,}056 (Pi~5) & Battery life & $\sim$\SI{4}{\hour} \\

\bottomrule
\end{tabular}

\vspace{2pt}
\end{table}

\begin{figure}
\centering
\includegraphics[width=0.88\linewidth]{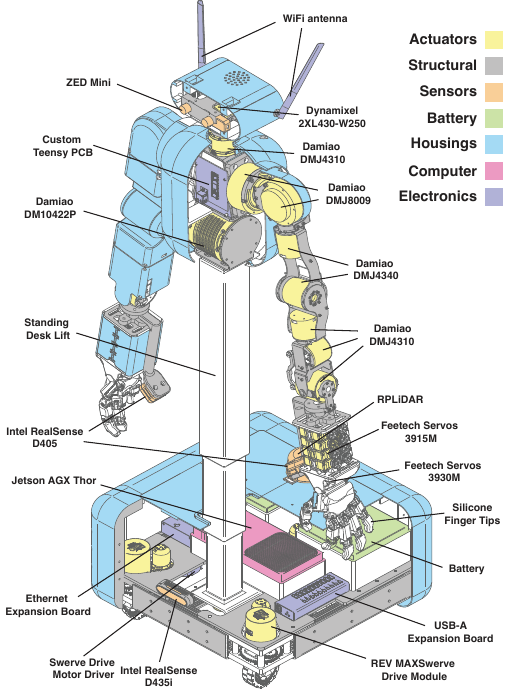}
\caption{Annotated cutaway of MABEL, showing actuators, sensors, compute, power, and electronics; colors group components by function.}
\label{fig:hardware_cutaway}
\end{figure}

\subsection{Mechanical Design}
MABEL uses three REV MAXSwerve modules instead of the four-module configurations in~\cite{yor,tidybot2024}.
Three non-collinear swerve drives provide holonomic motion with statically stable support, reducing the risk of wheel lift-off.
This configuration simplifies the drivetrain and saves about US\$700, at the cost of a smaller support polygon.
To mitigate the increased tip-over risk from the smaller support area, we introduce height-aware motion limits that scale base motion with lift height (\autoref{sec:motion}).
Two 7-DOF arms adapted from OpenArm~\cite{openarm} form the bimanual pair, allowing labs with existing OpenArm hardware to upgrade without replacing their arms.
Redesigned housings and wiring improve mechanical integration, and direct sourcing of components and actuators reduces the parts cost of both arms to approximately US\$3{,}000.
Each arm carries a 17-DOF tendon-driven ORCA hand~\cite{orcahand} through a modular wrist interface that also accepts other hands or grippers, such as the OpenArm gripper.
A quasi-direct-drive torso provides $-60^\circ$ to $15^\circ$ of pitch for reaching into low and deep workspaces.
The arms and torso are fully back-drivable, enabling low-impedance control and compliant physical interaction (\autoref{sec:wbc}).
Following YOR~\cite{yor}, a single FlexiSpot E7 standing-desk lift column is integrated to provide an additional \SI{0.635}{m} of vertical travel.
Together, the torso and lift extend the kinematic fingertip reach from floor level up to \SI{2.23}{m} (\autoref{fig:capability_cost_reach}).
A 3-DOF active neck carries the stereo head camera and lets the head look independently of the body for active perception, while torso pitch further extends downward visibility.
Neck and torso motion together roughly triple the camera coverage relative to a torso-fixed camera.
The body is proportioned to the operator to keep the embodiment gap small (\autoref{sec:retarget}): shoulder width, arm reach, and camera height follow adult reference dimensions~\cite{ansur2}, the \SI{63}{mm} stereo baseline matches human interpupillary distance~\cite{dodgson2004ipd}, and the neck and torso ranges cover typical human cervical and trunk motion~\cite{ferrario2002cervical}.
Each subsystem attaches through its own interface, so the body parts are modular: hands can be swapped for grippers or the upper body can be mounted on a fixed stand (\autoref{fig:hardware_cutaway}).

\subsection{Electrical, Sensors and Compute}
A custom modular Teensy-based board distributes power and handles data communication.
The same board serves both the mobile base and the upper body, each on its own controller area network (CAN) to distribute bandwidth and keep the system modular.
Each board supports up to three CAN buses and transistor--transistor logic (TTL) serial interfaces, and exchanges telemetry and commands with the host via Ethernet.
The embedded controllers handle low-level communication and actuator feedback, while the host runs ROS~2, teleoperation, and policy inference.
The software runs on Ubuntu and macOS and is validated on MacBook Pro (M1), Raspberry~Pi~5, Jetson Orin, and Jetson Thor.
Power comes from either a \SI{24}{V} battery pack, providing approximately \SI{4}{h} of operation, or swappable \SI{40}{V} power-tool batteries through a regulated \SI{24}{V} bus, providing approximately \SI{2}{h}; an external wall supply allows continuous use.
A synchronized stereo head camera and two wrist cameras provide the visual feedback for teleoperation and the observations for learning.
The experiments use a ZED Mini on the head and a global-shutter Intel RealSense D405 on each wrist, with \ang{102} and \ang{87} horizontal FOV, respectively.
Wide-angle Universal Serial Bus (USB) cameras with \ang{120} head and \ang{170} wrist FOV have also been tested as lower-cost replacements.
An optional base-mounted color-and-depth (RGB-D) camera and a light detection and ranging (LiDAR) sensor support localization and navigation.

\subsection{Fabrication and Replication}
\label{sec:replication}
MABEL is built primarily from off-the-shelf parts, with custom laser-cut sheet metal and 3D-printed components.
The Raspberry~Pi~5 configuration has a listed parts subtotal of \$8{,}056, excluding optional sensors.
We assembled a second unit from the released design and bill of materials in about two weeks, from parts to a fully tested robot.
All design files (STEP format), fabrication files, electronics, firmware, ROS~2 software, and simulation models are released, with original hardware under CERN-OHL-S-2.0 and software under Apache-2.0.

\section{Whole-Body Teleoperation and Control}
\label{sec:teleop}
\label{sec:control}\label{sec:onecontract}\label{sec:overview}\label{sec:wbc-notation}

\textbf{Notation.}
Operator frames are the headset's gravity-aligned world $\mathcal F_w$, head $\mathcal F_h$, wrists $\mathcal F_i$, and estimated body heading $\mathcal F_b$.
Robot frames are the torso $\mathcal F_r$, head camera $\mathcal F_g$, palms $\mathcal F_{e_i}$, and odometry $\mathcal F_o$, with $i\in\{\mathrm L,\mathrm R\}$ (\autoref{fig:avatar_teleop}(a)).
${}^{x}\Matrix R_y$, ${}^{x}\Vector p_y$ give the pose of $\mathcal F_y$ in $\mathcal F_x$; $\Matrix R_z(\psi)$ rotates by $\psi$ about the vertical.
Superscript $*$ marks a reference, subscript $a$ an anchor-latched value, and a bar an operator quantity scaled to the robot.
$\Vector q$ are joint positions with limits $[\Vector q_{\min},\Vector q_{\max}]$, $w$ cost weights, and $\Vector\nu_b=(\Vector v,\omega)$ the base twist in the robot base frame.

\subsection{Teleoperation Device and Networking}
\begin{figure*}[hpt!]
\centering
\includegraphics[width=\textwidth]{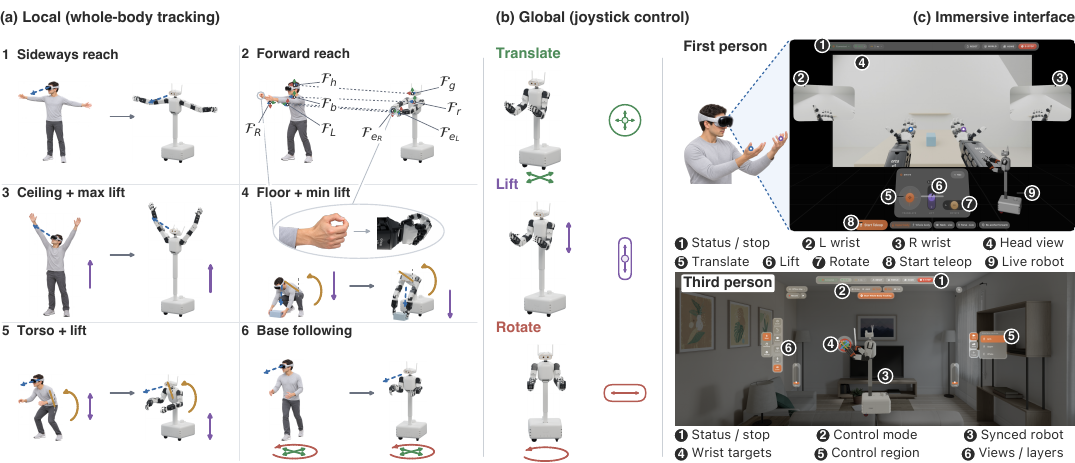}
\caption{(a) Operator--robot pose correspondence, labeled frames, and magnified hand detail.
(b) Joystick translation, lift, and yaw.
(c) First- and third-person interfaces with numbered controls; the third-person background is dimmed.
Arrows indicate head direction, torso pitch, lift, and base motion.}
\label{fig:avatar_teleop}
\end{figure*}

We use the Apple Vision Pro for its high-resolution displays, about 23 million pixels~\cite{apple2024visionprospecs}, and built-in head and hand tracking, which removes the need for controllers~\cite{hu2025xrrealitycheck,apple2024arkitwwdc}.
As in prior systems~\cite{park2024visionproteleop,opentelevision2024,bunnyvisionpro2024}, the operator sees through the robot's eyes (\autoref{fig:avatar_teleop}): each eye shows the head stereo camera image at true angular scale, and the \SI{63}{mm} camera baseline matches human interpupillary distance~\cite{dodgson2004ipd}, giving natural depth.
Head and hand poses stream at \SI{60}{Hz} to the robot for retargeting (\autoref{sec:armobj}).
In local mode, the whole body mirrors the operator, while in global mode, on-screen sticks drive the base and lift so the operator need not walk.
iPhone, browser, and Pico headset clients share the same interface.

\label{sec:transport}\label{sec:server}\label{sec:wire}\label{sec:client}
Commands use newest-only WebSocket delivery, so stale commands are dropped rather than queued, and video uses latest-frame queues over the User Datagram Protocol (UDP).
The headset connects over the robot's own Wi-Fi hotspot, a local area network (LAN), or Tailscale (\autoref{sec:exp-latency}).

\subsection{Retargeting}
\label{sec:armobj}\label{sec:retarget}\label{sec:retarget-safe}\label{sec:modes}
The Vision Pro tracks only the head, wrists, and 27 keypoints per hand, while MABEL has 56 DOF, so retargeting is ill-posed~\cite{avatarposer2022,egoposer2024}.
We estimate a reduced operator state and map each part to the joints that reproduce it.

\textit{Operator state.}
A calibration of approximately \SI{10}{s} estimates the operator's arm reach ($L_i^h$), head-to-shoulder offsets $({}^{b}\Vector d_i)$, and hand size.
Since torso orientation is not directly tracked, we approximate its yaw ($\psi$) using the common heading information inferred from the tracked head and wrist orientations, and define the gravity-aligned body-heading frame by ${}^{w}\Matrix R_b=\Matrix R_z(\psi)$.
A comfort box $\mathcal B=[-0.02,0.18]\times[-0.12,0.12]$\,m splits the head displacement into lean and walking:
\begin{equation}
\label{eq:headdisp}
\begin{aligned}
\begin{bmatrix}\Delta\Vector r\\ \Delta z\end{bmatrix}&={}^{w}\Matrix R_b^\top({}^{w}\Vector p_h-{}^{w}\Vector p_{h,a}),\\[2pt]
\Delta\Vector r_{\rm lean}&=\operatorname{clip}_{\mathcal B}(\Delta\Vector r),\quad
\Delta\Vector r_{\rm walk}=\Delta\Vector r-\Delta\Vector r_{\rm lean},
\end{aligned}
\end{equation}
where $\Delta\Vector r=[\Delta x,\Delta y]^\top$ is forward and lateral displacement, $\Delta z$ vertical displacement, and $\operatorname{clip}_{\mathcal B}$ projects onto $\mathcal B$.
The bounded component $\Delta\Vector r_{\rm lean}$ provides a leaning reference, while the residual $\Delta\Vector r_{\rm walk}$ represents walking intent.
Anchors reset on recentering, clutch release, and tracking loss to bound drift in $\psi$.

\textit{Body.}
The operator's forward leaning ($\Delta x_{\rm lean}$, the first component of $\Delta\Vector r_{\rm lean}$), vertical displacement ($\Delta z$), and walking displacement ($\Delta\Vector r_{\rm walk}$) are mapped to the robot's desired torso pitch ($q_{\rm torso}^*$), lift height ($z_{\rm lift}^*$), and base position (${}^{o}\Vector r_{\rm base}^*$), respectively:
\begin{equation}
\label{eq:body}
\begin{aligned}
q_{\rm torso}^*&=q_{\rm torso,0}-\gamma\,\Delta x_{\rm lean},\\[2pt]
z_{\rm lift}^*&=z_{\rm lift,a}+\Delta z,\\[2pt]
{}^{o}\Vector r_{\rm base}^*&={}^{o}\Vector r_{\rm base,a}+\Matrix R_z(\psi_{o,a})\,\Delta\Vector r_{\rm walk},
\end{aligned}
\end{equation}
where $q_{\rm torso,0}$ is the upright torso pitch, $\gamma$ the displacement-to-pitch gain, and $\psi_{o,a}$ the robot heading in $\mathcal F_o$ at the anchor.

\textit{Neck.}
The desired neck joint angles ($\Vector q_{\rm neck}^*$) reproduce the operator's head orientation (${}^{w}\Matrix R_h$) relative to the estimated body-heading frame (${}^{w}\Matrix R_b$):
\begin{equation}
\label{eq:neck}
\Vector q_{\rm neck}^*=\operatorname{ypr}\bigl({}^{w}\Matrix R_b^\top\,{}^{w}\Matrix R_h\bigr)-\begin{bmatrix}0 & q^*_{\rm torso} & 0\end{bmatrix}^\top,
\end{equation}
where $\operatorname{ypr}(\cdot)$ returns yaw, pitch, and roll for the neck's joints.

\textit{Arms.}
Each wrist keeps its offset from the shoulder, scaled by the reach ratio $s_i=L_i^r/L_i^h$ of robot to operator arm reach~\cite{gleicher1998,gmr2026}:
\begin{equation}
\label{eq:wristmap}
{}^{r}\Vector p_{e_i}^*={}^{r}\Vector o_{r, i}+s_i\,{}^{w}\Matrix R_b^\top({}^{w}\Vector p_i-{}^{w}\Vector o_{h, i}),
\end{equation}
where ${}^{w}\Vector o_{h,i}={}^{w}\Vector p_h+{}^{w}\Matrix R_b\,{}^{b}\Vector d_i$ and ${}^{r}\Vector o_{r,i}$ are the operator and robot shoulders, respectively, and the axes of $\mathcal F_b$ are identified with those of $\mathcal F_r$.
The inverse kinematics (IK) problem for the seven arm joints and the hand's wrist joint, $\Vector q\in\mathbb R^8$,
is formulated as a nonlinear least-squares optimization:
\begin{equation}
\label{eq:armobj}
\begin{aligned}
\Vector q_{\rm IK}=\arg\min\nolimits_{\Vector q}\;\,
&\|{}^{r}\Vector p_{e_i}(\Vector q)-{}^{r}\Vector p_{e_i}^*\|^2\\[2pt]
&+w_{\rm dir}\textstyle\sum_{j\in\mathcal U}
\bigl(1-\Vector u_j(\Vector q)^\top\Vector u_j^*\bigr)\\[2pt]
&+w_{\rm rest}\|\Vector q-\Vector q_{\rm rest}\|^2\\[2pt]
\text{s.t.}\;\,
&\Vector q_{\min}\le\Vector q\le\Vector q_{\max},
\end{aligned}
\end{equation}
where $\mathcal U$ indexes the palm normal, finger axis, and elbow direction, with targets derived from ${}^{w}\Matrix R_b^\top{}^{w}\Matrix R_i$ and an elbow prior.
The weights $w_{\rm dir}$ and $w_{\rm rest}$ keep the problem well-conditioned and bias solutions toward the straight-arm rest posture $\Vector q_{\rm rest}$.
The problem is solved by the damped Gauss--Newton method, warm-started from the previous solution.

\textit{Hands.}
The 16 finger joints $\Vector q_{\rm f}$ of each ORCA hand are retargeted to match the operator's hand fingertip positions in the palm frame~\cite{dexpilot2020,anyteleop2023}:
\begin{equation}
\label{eq:finger}
\begin{aligned}
\Vector q_{\rm f}^*=\arg\min\nolimits_{\Vector q_{\rm f}}\;\,
&\|\Matrix P(\Vector q_{\rm f})-\bar{\Matrix P}\|_F^2\\[2pt]
&+w_{\rm pinch}\textstyle\sum_{m}\|\Vector\delta_m(\Vector q_{\rm f})-\bar{\Vector\delta}_m\|^2\\[2pt]
\text{s.t.}\;\,&\Vector q_{\rm f,\min}\le\Vector q_{\rm f}\le\Vector q_{\rm f,\max},
\end{aligned}
\end{equation}
where $\Matrix P$ stacks the robot fingertip positions and $\Vector\delta_m$ is the vector from the thumb to fingertip $m$, both in the palm frame; the barred $\bar{\Matrix P}$ and $\bar{\Vector\delta}_m$ are the operator's counterparts, scaled to the robot hand by the calibrated hand-size ratio.
The weight $w_{\rm pinch}$ emphasizes the thumb-to-fingertip vectors, which preserves pinch apertures during precise grasping.

\subsection{Motion Model}
\label{sec:motion}\label{sec:tipcond}\label{sec:comh}\label{sec:proj}
\textit{Swerve kinematics.}
For a base twist $(\Vector v,\omega)$ in the base frame, module $j\in\{1,2,3\}$ at position $\Vector r_j$ from the center of rotation moves with velocity
\begin{equation}
\label{eq:swerve}
\Vector v_j=\Vector v+\omega\,\Matrix S\Vector r_j,\qquad
\Matrix S=\begin{bmatrix}0&-1\\1&0\end{bmatrix},
\end{equation}
where $\Matrix S$ is a $90^\circ$ rotation, so $\omega\Matrix S\Vector r_j$ is the planar cross product.
The direction and magnitude of $\Vector v_j$ set the module's steering angle and drive speed, with drive speed scaled by $\cos e_j$, where $e_j$ is the steering error, so the base moves in any direction once re-steered.

\textit{Height-aware limits.}
The three-module base has a smaller support polygon, so base motion is limited by lift height and commanded velocity through a center-of-mass (CoM) tip-over model~\cite{sugano1993}.
For a fixed planar CoM offset $\Vector r_c$, the horizontal acceleration is
\begin{equation}
\label{eq:accel}
\Vector a_c\approx
\underbrace{\dot{\Vector v}+\omega\Matrix S\Vector v}_{\text{base}}
+\underbrace{\dot\omega\Matrix S\Vector r_c}_{\text{tangential}}
\;\underbrace{-\,\omega^2\Vector r_c}_{\text{centripetal}},
\end{equation}
neglecting CoM motion from upper-body posture, with CoM height $h_c(z_{\mathrm{lift}})=h_0+k_h z_{\mathrm{lift}}$ fitted from the robot model.
The zero-moment-point (ZMP) approximation~\cite{vukobratovic2004zmp} gives edge-wise bounds and a utilization
\begin{equation}
\label{eq:zmp}
a_k^{\max}=\frac{g\,\ell_k}{\sigma_k\,h_c(z_{\mathrm{lift}})},\qquad
u=\max_k\frac{-\Vector n_k^\top\Vector a_c}{a_k^{\max}},
\end{equation}
where $\Vector n_k$ and $\ell_k$ are the outward normal and CoM distance of edge $k$, $\sigma_k\ge1$ a safety factor, and $g$ gravity.
Base commands are scaled to keep $u\le1$, slowing the base as the lift rises.

\subsection{Whole-Body Controller}
\label{sec:wbc}\label{sec:plant}\label{sec:nested}\label{sec:unify}\label{sec:wrench}\label{sec:hwrender}
\textit{Collision avoidance.}
The arm solutions $\Vector q_{\rm IK}\in\mathbb R^{16}$ are projected onto the closest safe joint velocity~\cite{faverjon1987,ames2019}:
\begin{equation}
\label{eq:qp}
\begin{aligned}
\dot{\Vector q}^*=\arg\min\nolimits_{\dot{\Vector q}}\;\,
&\tfrac12\|\dot{\Vector q}-\dot{\Vector q}_{\rm des}\|^2\\[2pt]
\text{s.t.}\;\,&\dot{\Vector q}_{\min}\le\dot{\Vector q}\le\dot{\Vector q}_{\max},\\[2pt]
&-\Vector J_c\dot{\Vector q}\le\kappa\,(d_c-d_s),\quad\forall c\in\mathcal C,
\end{aligned}
\end{equation}
where $\dot{\Vector q}_{\rm des}=(\Vector q_{\rm IK}-\Vector q)/\Delta t$ with control period $\Delta t$, the bounds combine velocity and joint limits, and $\Vector q^*=\Vector q+\Delta t\,\dot{\Vector q}^*$ is the reference for \eqref{eq:imp}.
For each link-capsule pair $c\in\mathcal C$, $d_c$ is the clearance, $\Vector J_c=\partial d_c/\partial\Vector q$, $d_s$ a margin, and $\kappa$ a gain.

\begin{figure*}[t]
\centering
\definecolor{forceTeal}{HTML}{3A9A94}
\includegraphics[width=\textwidth]{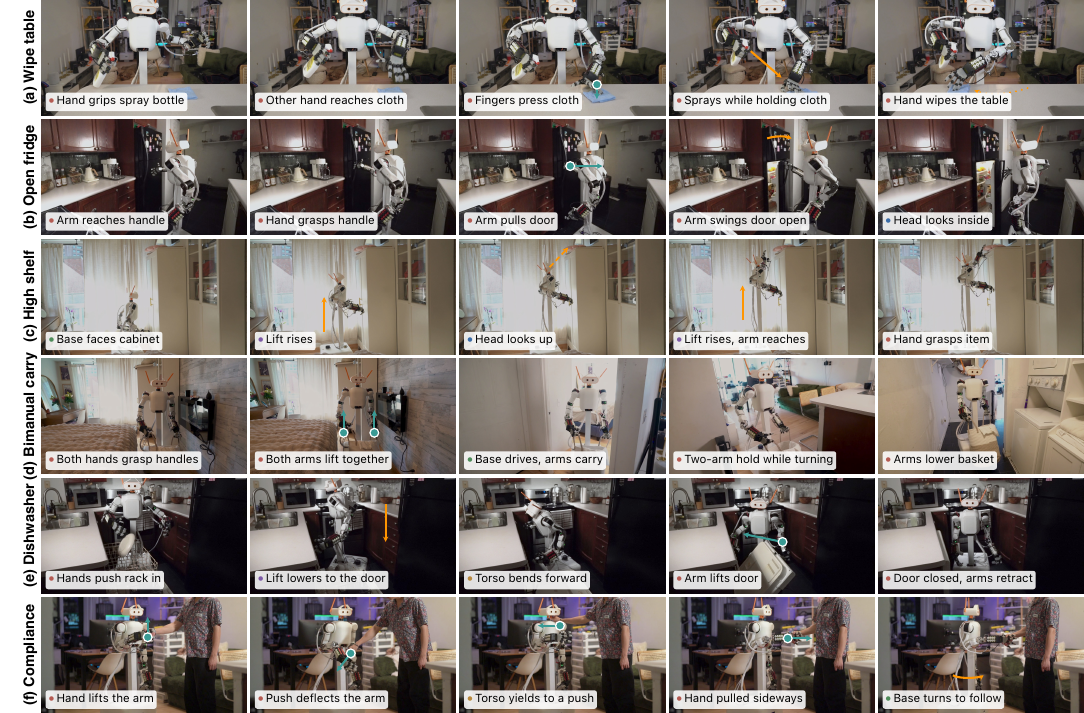}
\caption{Teleoperated household tasks on the physical robot, one continuous trial per row, time left to right.
Arrows mark lift and door motion, the dashed arrow the gaze, the dotted path the hand, and \textcolor{forceTeal}{teal} arrows contact forces. Each label names the body part the step relies on:
\textcolor[HTML]{3B6CA8}{$\bullet$}\,head, \textcolor[HTML]{B8893E}{$\bullet$}\,torso, \textcolor[HTML]{7C5AA6}{$\bullet$}\,lift, \textcolor[HTML]{4E8C5A}{$\bullet$}\,base, and \textcolor[HTML]{B45B52}{$\bullet$}\,arms and hands.
Row (f) shows compliance: a person lifts, pushes, and pulls the robot by hand (\autoref{sec:wbc}).}
\label{fig:household_tasks}
\end{figure*}

\textit{Compliant control.}
The arms and torso track joint references using joint-space impedance control~\cite{hogan1985}:
\begin{equation}
\label{eq:imp}
\begin{aligned}
\Vector{\tau}
={}&
\Vector{\tau}_{\mathrm{g}}(\Vector{q})
+\Vector{\tau}_{\mathrm{f}}(\dot{\Vector{q}})+
\Matrix{K}(\Vector{q}^{*}-\Vector{q})
+\Matrix{D}(\dot{\Vector q}^*-\dot{\Vector q}),
\end{aligned}
\end{equation}
where $\Vector{\tau}_{\mathrm{g}}$ and $\Vector{\tau}_{\mathrm{f}}$ compensate gravity and friction, and $\Matrix{K}$ and $\Matrix{D}$ set joint stiffness and damping.
Because system identification keeps model errors small, $\Matrix{K}$ and $\Matrix{D}$ can be set low so that the arms and torso yield under external contact.
Physical demonstrations show manual displacement of the arms and torso, with the arm returning toward its reference after release (\autoref{fig:household_tasks}, row~f).
The hands, neck, and lift use position control.

\section{Simulation and Learning Infrastructure}
\label{sec:autonomy}\label{sec:dataengine}\label{sec:studios}

\subsection{Simulation}
\label{sec:simulation}
We provide full-body MuJoCo and Isaac Lab models that include all 59 actuated joints, together with 35 pre-built static and dynamic scenes.
The digital twin also simulates camera and LiDAR streams for data collection and policy learning.
A simulation bridge exposes the same command interface as the hardware, allowing teleoperation and policy execution to use the same controller and retargeter in simulation and on hardware.

\subsection{Data Collection}
\label{sec:data_collection}
\begin{figure}[ht]
\centering
\includegraphics[width=\columnwidth]{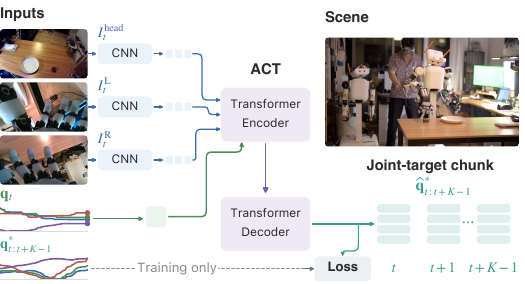}
\caption{ACT learning pipeline.
Head and wrist images $I_t$ and measured joint states $\Vector q_t$ condition the prediction of a joint-reference chunk $\widehat{\Vector q}_{t:t+K-1}^*$, supervised during training by the demonstrated references $\Vector q_{t:t+K-1}^*$.
The scene shows the pick-and-place task used for data collection and deployment.}
\label{fig:learning_demonstration}
\end{figure}

We collect Vision Pro demonstrations of grasping a small object with the ORCA dexterous hands and placing it on a plate (\autoref{fig:learning_demonstration}).
We use a 59-coordinate joint representation $\Vector q_t\in\mathbb{R}^{59}$: 14 arm joints, 34 hand joints (16 finger joints and one wrist per hand), three neck joints, the torso, one lift height, and six swerve steering/drive coordinates.
The two coupled lift stages are combined into one height; the hand coordinates already include both wrists.
Each episode retains stereo head RGB and both wrist RGB streams at a nominal \SI{30}{Hz}, available depth, and timestamped joint feedback, commanded references $\Vector q_t^*$, base twist $\Vector\nu_{b,t}$, and lift-jog commands at a nominal \SI{50}{Hz}.
Base-twist and lift-jog commands are stored separately from the 59 joint coordinates.
Raw camera and control records retain their timestamps and validity information; curated episodes are exported to LeRobot v3~\cite{lerobot} on a \SI{30}{Hz} grid using the latest available sample from each stream.

\subsection{Robot Learning}
We train an ACT policy~\cite{aloha2023} using the LeRobot framework.
Each camera image is encoded separately and fused with the 59-coordinate state $\Vector q_t$ to predict a $K$-step joint-reference chunk $\widehat{\Vector q}_{t:t+K-1}^*$, including both hands' finger and wrist references.
The demonstrated references $\Vector q_{t:t+K-1}^*$ supervise training; the hat denotes a prediction.
The predicted references are executed through the same whole-body controller used for teleoperation (\autoref{sec:wbc}); results are reported in \autoref{sec:exp-learning}.

\section{Experiments and Results}
\label{sec:experiments}\label{sec:exp-platform}

\subsection{Repeatability and Path Following}
\label{sec:exp-repeat}\label{sec:exp-path}
\begin{figure}[ht]
\centering
\includegraphics[width=\columnwidth]{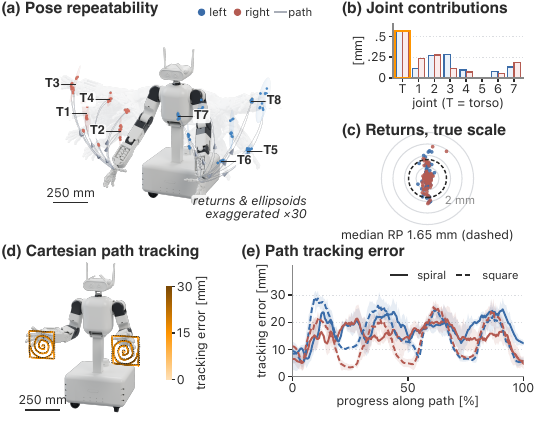}
\caption{Hardware repeatability and path following illustration and results.
(a) Ten cycles, with return samples and $95\%$ covariance ellipsoids magnified $30\times$.
(b) Joint contributions to scatter.
(c) True-scale returns; \SI{1}{mm} rings, dashed median repeatability (RP).
(d) Twelve measured traces; dashed references.
(e) Tracking error: mean and range over three trials.}
\label{fig:repeatability_path_following}
\end{figure}

To test the hardware capabilities, we evaluate upper-body repeatability and Cartesian tracking under proportional--derivative (PD) control with gravity compensation (\autoref{fig:repeatability_path_following}).
For the repeatability test, each palm visits four different checkpoints from a common home pose over 10 cycles.
The resulting median repeatability is \SI{1.65}{mm}, with a best case of \SI{0.46}{mm}.
The torso contributes the largest joint-induced error (\SI{0.56}{mm} root mean square (RMS) per palm; \autoref{fig:repeatability_path_following}(b)).
For path following, both palms trace mirrored \SI{10}{cm}-radius spirals and \SI{20}{cm} squares over three trials at torso pitches of $0^\circ$ and $-10^\circ$.
Geometric path error ranges from \SIrange{3.7}{6.7}{mm} RMS, while time-aligned tracking error reaches \SIrange{15}{19}{mm} RMS because the palm lags the reference.
Compensating for this lag reduces tracking error to \SIrange{5}{9}{mm} RMS.
Gravity feedforward had the largest effect, reducing palm error from \SI{194}{mm} to \SI{45}{mm} under compliant gains.

\subsection{Wheel Odometry}
\label{sec:exp-wheel-odometry}
\begin{figure}[ht]
\centering
\includegraphics[width=\columnwidth]{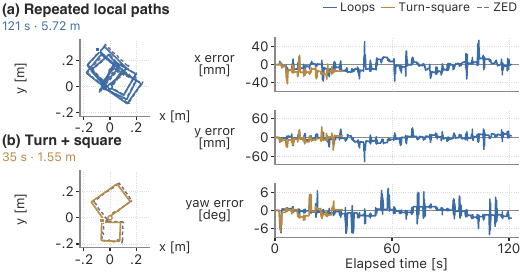}
\caption{Wheel odometry illustration: trajectories and signed x/y/yaw disagreements versus reference trajectory.}
\label{fig:base_odometry}
\end{figure}

Wheel odometry integrates drive-encoder changes and steering angle from the three swerve modules, and is compared against the ZED camera's visual-inertial position and inertial measurement unit (IMU) heading.
Over a repeated local path (\SI{5.72}{m} in \SI{121}{s}) and a turn-and-square trial (\SI{1.55}{m} in \SI{35}{s}), the RMS disagreement is \SI{13.5}{mm} and \SI{17.7}{mm} in position and $1.63^\circ$ and $1.47^\circ$ in yaw (\autoref{fig:base_odometry}).
The base references of \eqref{eq:body} are tracked in the odometry frame, for which this centimeter-level accuracy is sufficient over a teleoperation session.

\subsection{Teleoperation Latency}
\label{sec:exp-teleop}\label{sec:exp-operator}\label{sec:exp-latency}
\begin{figure}[ht]
\centering
\includegraphics[width=\columnwidth]{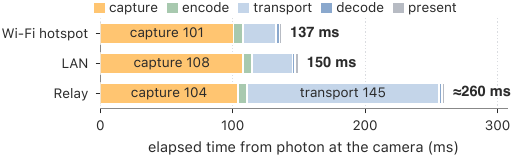}
\caption{Glass-to-glass video latency per connection method, from camera photon to operator pixel.
Hotspot and LAN totals are measured; the relay total ($\approx$) is partly derived from measured encode and transport times.}
\label{fig:video_latency}
\end{figure}
We measure glass-to-glass video latency by filming a Gray-coded clock on the operator's screen with the head camera and timing when each frame reappears on the same screen, so both ends share one clock (\autoref{fig:video_latency}).
Latency is $137\pm5$~ms over the robot's hotspot, $150\pm0.2$~ms over the LAN, and $161\pm16$~ms over direct peer-to-peer Tailscale (mean $\pm$ standard deviation of per-trial medians, three \SI{20}{s} trials; four for Tailscale).
The camera dominates, contributing 101--108~ms on every route, while encoding, decoding, and display add about \SI{12}{ms}.
Transport is what varies: \SI{25}{ms} on the hotspot, \SI{31}{ms} on the LAN, and \SI{145}{ms} through the relay, for a relay total of about \SI{260}{ms} through our public cloud server (Tailscale is not plotted).

\subsection{Policy Learning}
\label{sec:exp-learning}
We evaluate the ACT policy of \autoref{sec:autonomy} on grasping a small object and placing it on a plate with both ORCA hands installed (\autoref{fig:learning_demonstration}).
We collected 80 whole-body Vision Pro demonstration takes, segmented into 117 episodes totaling \SI{38.0}{min} (68{,}364 frames) at \SI{30}{Hz}.
The demonstrations are mostly upper-body, with the base and lift near their initial configuration, so this experiment validates the pipeline rather than mobile behavior; the policy nonetheless predicts the full joint-reference representation of \autoref{sec:data_collection} and executes through the same whole-body controller as teleoperation.
Inputs are the left head image and both wrist images at $320\times240$ and the past \SI{1}{s} of joint positions sampled at \SI{5}{Hz}; the policy predicts $K=120$ references (\SI{4}{s}) at \SI{30}{Hz}, of which the first \SI{1}{s} is executed before re-planning, without temporal ensembling.
We split by take (64/8/8 train/val/test) and train for 8{,}000 steps with batch size 24 on an Apple M1 Max in 6\,h.
Over 10 autonomous trials with the object placed randomly within a $20\times20$\,cm region and a \SI{60}{s} limit, the policy succeeds in 8.

\section{Limitations and Future Work}
The current lift provides a large vertical workspace but trades speed for affordability.
Future designs may adopt a faster, folding body with additional waist, knee, and ankle joints replacing the lift for more natural whole-body motion.
The OpenArm's link proportions also differ from human anatomy, motivating more anthropomorphic arms to improve reach and retargeting.
Future work will extend MABEL toward long-horizon autonomy through longer-horizon whole-body policies, onboard localization and mapping, and foundation-model-based task planning over learned skills.

\section{Conclusion}
\label{sec:conclusion}
We presented MABEL, an open-source 56-DOF mobile bimanual platform that combines mobility, a large workspace, active perception, dexterity, and compliant actuation at roughly a tenth of the cost of comparable commercial systems.
Immersive whole-body teleoperation and a shared control and learning stack turn that hardware into a complete pipeline from demonstration to deployed policy.
By releasing the complete design and software, MABEL lowers the barrier to research in mobile manipulation and robot learning.
Much as Linux gave a community a shared open system to build on, we hope MABEL becomes a platform the robotics community develops together.

\bibliographystyle{IEEEtran}
\bibliography{citations}

\end{document}